\documentclass[11pt]{article}

\usepackage[final]{acl}

\usepackage{times}
\usepackage{latexsym}

\usepackage[T1]{fontenc}

\usepackage[utf8]{inputenc}
\NewDocumentCommand{\heba}
{ mO{} }{\textcolor{magenta}{\textsuperscript{\textit{heba}}\textsf{\textbf{\small[#1]}}}}
\NewDocumentCommand{\kevin}
{ mO{} }{\textcolor{blue}{\textsuperscript{\textit{kevin}}\textsf{\textbf{\small[#1]}}}}
\NewDocumentCommand{\peter}
{ mO{} }{\textcolor{red}{\textsuperscript{\textit{peter}}\textsf{\textbf{\small[#1]}}}}
\NewDocumentCommand{\giuseppe}
{ mO{} }{\textcolor{brown}{\textsuperscript{\textit{giuseppe}}\textsf{\textbf{\small[#1]}}}}
\NewDocumentCommand{\yilun}
{ mO{} }{\textcolor{cyan}{\textsuperscript{\textit{yilun}}\textsf{\textbf{\small[#1]}}}}

\usepackage{microtype}

\usepackage{inconsolata}

\usepackage{graphicx}
\usepackage[para]{threeparttable}
\usepackage{booktabs}
\usepackage{multirow}
\usepackage{graphicx}
\usepackage{amsmath}
\usepackage{amssymb}
\usepackage{pifont}
\usepackage{hyperref}
\usepackage{mathtools}
\usepackage{array}
\usepackage{colortbl}
\usepackage[ruled,vlined]{algorithm2e}
\usepackage{tabularx}
\usepackage[dvipsnames]{xcolor}
\usepackage{tcolorbox}
\usepackage{enumitem}
\usepackage[textsize=scriptsize]{todonotes}
\newcommand{\xmark}{\ding{55}} 
\newcommand{\cmark}{\ding{51}}
\usepackage{float}

\newcommand{\topiocqa}{TopiOCQA\xspace}
\newcommand{\qrecc}{QReCC\xspace}

\newcommand{\golddoctest}{Identifying Ground-truth Documents\xspace}
\newcommand{\randdoctest}{Less Relevance But More Utility\xspace}
\newcommand{\samedoctest}{Same Document But Different Utility\xspace}
\newcommand{\wrand}{`w/ rand'\xspace}
\newcommand{\wretrieve}{`w/ retr'\xspace}
\newcommand{\grogu}{\textsc{GroGU}\xspace}

\newcommand{\aautoref}[1]{\hyperref[#1]{Appendix~\ref*{#1}}}
\title{Evaluating the Utility of Grounding Documents\\with Reference-Free LLM-based Metrics}

\author{Yilun Hua$^{\dagger*}$, Giuseppe Castellucci$^\ddagger$, Peter Schulam$^\ddagger$,
\textbf{Heba Elfardy}$^\ddagger$, \textbf{Kevin Small}$^\ddagger$ \\
  $^\dagger$Department of Computer Science and Cornell Tech, Cornell University\\ 
  $^\ddagger$Amazon \\
  yilunhua@cs.cornell.edu \\
  \{giusecas,schulamp,helfardy,smakevin\}@amazon.com}

\begin{document}
\maketitle
\renewcommand*{\thefootnote}{*}
\footnotetext{Work done as an intern at Amazon.}
\renewcommand*{\thefootnote}{\arabic{footnote}} 
\begin{abstract}
Retrieval Augmented Generation (RAG)'s success depends on the \emph{utility} the LLM derives from the content used for grounding. Quantifying content utility does not have a definitive specification and existing metrics ignore model-specific capabilities and/or rely on costly annotations. In this paper, we propose {\em Grounding Generation Utility} (\grogu), a model-specific and reference-free metric that defines utility as a function of the downstream LLM's generation confidence based on entropy. Despite having no annotation requirements, \grogu is largely faithful in distinguishing ground-truth documents while capturing nuances ignored by LLM-agnostic metrics. We apply \grogu to train a query-rewriter for RAG by identifying high-utility preference data for Direct Preference Optimization. Experiments show improvements by up to 18.2 points in Mean Reciprocal Rank and up to 9.4 points in answer accuracy.

\end{abstract}

\section{Introduction}
Retrieval Augmented Generation (RAG) is the dominant approach for providing Large Language Models (LLMs) with knowledge not present in their training data. The basic approach to RAG is straightforward: the user question is sent to a \emph{retriever} that returns documents from an index. The documents are then combined with the original input and sent to a downstream LLM (the \emph{generator}) to generate a response. In practice, however, a RAG system typically has many sub-components that must be carefully tuned to achieve strong performance. For example, a \emph{query-rewriter}~\citep{moConvGQRGenerativeQuery2023} is often needed to reformulate a question in a conversational interaction, which otherwise omits key information (example in \aautoref{append:illustration}). The reformulated queries retrieve more useful grounding documents for the downstream LLM, thus improving the corresponding response quality.

Tuning a RAG system is nontrivial and typically involves making changes to individual RAG components. In practice, such tuning is usually guided by metrics that rely on annotated labels as references. Examples include the rank of gold passages among retrieved documents or comparing the generated answers with gold ones to compute accuracy. Reference annotations have been used to improve components such as the query-rewriter~\cite[e.g.,][]{yoonAskOptimalQuestions2025,laiAdaCQREnhancingQuery2025} or a re-ranker~\cite[e.g.,][]{sunDynamicRAGLeveragingOutputs2025}, to provide more \emph{useful} grounding documents to the generator. An important limitation of such previous works is the reliance on annotated labels, which limits the adoption of more automated approaches to tuning RAG systems and extending to new domains.

These limitations would be resolved if we can measure the utility of retrieved documents without requiring reference labels. One naive option is to use the \emph{relevance} scores assigned by a retriever, i.e., an LLM-agnostic relation between a query and a document.
In practice, such relevance scores are not used for model training due to reasons such as their noisiness (a higher-ranked document, i.e., the one assigned a higher relevance score by a retriever, may have less actual relevance than a lower-ranked one)~\citep{yu2024rankragunifyingcontextranking}. We elaborate on other practical issues in \autoref{sec:relevance_is_not_utility}. In this paper, we argue that relevance scores are inherently flawed as an estimation of the utility of documents in a RAG setting due to their LLM-agnostic nature. Even an ideal relevance scorer will ignore important nuances in how much utility an LLM can derive from the grounding documents.

We refer to this \emph{utility} of grounding documents as \emph{Grounding Generation Utility} (\grogu). In this paper, we define \grogu as a \emph{reference-free} metric that captures how useful the documents are to a specific LLM. We define \grogu based on the LLM's generation confidence and propose a specific formulation that modifies the standard token entropy calculation for more robust performance, without relying on any annotations. We show that our formulation can still capture the general notion of retrieval relevance but it additionally models aspects that go beyond relevance (e.g., when the same document has different utilities for different LLMs or when less relevant documents have more utility). Furthermore, we show how our \grogu metric can be used to improve existing components of a RAG system: as an example task, we show that \grogu can be used to identify preference data without any manual annotations for training a query-rewriter for RAG with Direct Preference Optimization (DPO)~\citep{rafailov2024directpreferenceoptimizationlanguage}. We test this method with both sparse and dense retrievers, showing consistent improvements in retrieval performance, by up to 18.2 points in Mean Reciprocal Rank (MRR) as well as in the downstream generator performance, with improvements by up to 9.4 percentage points in accuracy.\footnote{We will release our code upon acceptance.}

This paper proceeds as follows. We elaborate on the motivation and background of \grogu in Section~\ref{sec:relevance_is_not_utility}, define \grogu in Section \ref{sec:gu_definition}, and explore how it addresses limitations of relevance scores in Section \ref{sec:inference_only_tests}. In Section \ref{sec:training}, we use \grogu to train a query-rewriter. Finally, in Sections \ref{sec:related_and_future_work} and \ref{sec:conclusion} we discuss related works and draw conclusions, respectively.

\section{Motivation and Background}

\label{sec:relevance_is_not_utility}
Existing methods for tuning components of a RAG system primarily use annotated gold passages or gold answers as references. These required annotations are often costly and need to be re-executed for every domain. Moreover, this approach does not work well in domains where it is difficult to define a ``correct'' answer or a gold passage~\citep{ananthaOpenDomainQuestionAnswering2021}. One solution is to use a \emph{reference-free} metric, such as retriever relevance scores. However, these scores have not served as good training signals to improve RAG components (e.g., query-rewriter, reranker) due to multiple reasons.
First, retrievers are optimized for recall over precision in document relevance ranking and there exist challenges in learning effective local alignments across the entire embedding space to support accurate matching~\citep{luan-etal-2021-sparse,yu2024rankragunifyingcontextranking}. As a result, the retrievers are noisy: given any two documents, the one ranked higher by a retriever may actually have lower relevance than the other one~\citep{wang2023learningfiltercontextretrievalaugmented,yoran2024makingretrievalaugmentedlanguagemodels}. A second issue stems from the retriever's role in a RAG pipeline: (1) retriever's relevance scores are produced before the document reranking step, thus not reflecting the quality of the reranker or providing reward signals to improve the reranker and (2) for query-rewriter, relevance scores are produced based on rewritten queries but scores of different rewritten queries are not comparable. One that misinterprets the user intent (e.g., asks a different question) may still have documents of higher relevance in the corpus than a correct rewritten query.

Finally, in this paper, we further claim that relevance is fundamentally insufficient for measuring the usefulness of grounding documents because it is LLM-agnostic and thus does not consider how a particular document will be utilized by a particular LLM. We present scenarios where even an ideal relevance scorer would fail, demonstrating the need to distinguish between the generation utility of documents and their retrieval relevance.

\paragraph{\samedoctest:} An LLM-agnostic relevance scorer will assign the same score to a document irrespective of which LLM will utilize the document. However, different models often derive different levels of utility from the same document. \autoref{fig:same_doc_diff_utils} shows that two LLMs initially fail to answer a question (using parametric knowledge). When conditioning on the same document, only one of them gives the correct answer.  
We further show this phenomenon in \autoref{sec:samedoctest}.
\begin{figure}[t]
\centering
\setlength{\fboxsep}{4pt}
\setlength{\fboxrule}{0.5pt}
\fbox{%
\parbox{0.95\columnwidth}{%
\small
\textbf{Question:} Who lives in the blue house in Balamory? \\
\textbf{Answer:} Edie McCredie \\

\vspace{-5pt}
\textbf{Document:} \\
{\footnotesize
(Title: Balamory) Character Actor Job House Song (s) Miss Hoolie Julie Wilson Nimmo Nursery teacher Green houseNursery ...
\textcolor{ForestGreen}{Edie McCredie Juliet Cadzow Bus driver Blue garage / house}
When I Honk My Horn ( Series 1 - 4 ) Let Me TakeYou on a Journey ( Series 3 — 4 ) PC Plum Andrew Agnew ...
}\\

\vspace{-5pt}
\textbf{Qwen-2-1.5b-it (1.5B):}\\
{\footnotesize
Ungrounded: The character named ``Bella'' \textcolor{red}{\xmark}\\
Grounded: The Blue House is owned by Mr. and Mrs. Green \textcolor{red}{\xmark}
}\\
\vspace{-5pt}

\textbf{Phi-4 (14B):}\\
{\footnotesize
Ungrounded: The blue house is home to Mr. Bloom. \textcolor{red}{\xmark}\\
Grounded: Edie McCredie, played by Juliet Cadzow. \textcolor{ForestGreen}{\cmark}
}
}%
}
\caption{The same document may have different LLM-specific utility. The weaker model (Qwen-2-1.5-it) cannot utilize this document likely due to its challenging format. Both models are sampled 50 times under Qwen's default sampling configuration, with consistent results.}
\label{fig:same_doc_diff_utils}
\end{figure}

\paragraph{\randdoctest:} Recently, \citet{cuconasuPowerNoiseRedefining2024} reported the counterintuitive finding that adding irrelevant documents to an LLM's context improves the correctness of the final response (given the context also includes a document with the correct answer). Moreover, the degree to which noise improves response quality depends on the LLM used to generate the response. This result runs counter to the intuition of relevance scores, which always assigns low scores to random documents, while these random documents may have high generation utility. We show further evidence of this phenomenon in \autoref{sec:randdoctest}.

We address these issues by proposing Grounding Generation Utility (\grogu), which exploits the model's generation confidence without requiring the ground-truth answer. \grogu can significantly automate the process of fine-tuning a RAG system by avoiding human annotations while also capturing nuances that relevance scores typically ignore.

\section{Grounding Utility}
\label{sec:gu_definition}
We design {\em Grounding Generation Utility} (\grogu) using the LLM's generation confidence. In particular, we propose to measure the change in confidence of an LLM when conditioned (and not conditioned) on documents. We focus on the scenario where the ground-truth answer is unknown and evaluation must be done without any reference to ground-truth annotations.
Although this approach does not fully replace relevance scores or annotation-based evaluation, we demonstrate that it complements existing paradigms.

\subsection{Preliminaries}

Formally, we define \grogu such that it depends on the specific language model $\theta$, a question $q$, and a list of retrieved documents $\mathbf{D_r}$ placed in the context window of the model (or the ``grounding context'' for short). We want to quantify the utility of $\mathbf{D_r}$ for the model $\theta$ to generate a response to the question $q$, that is, $\text{\grogu}_\theta(q,\mathbf{D_r})$. 

\subsection{Confidence Difference as Utility}
We define the utility of the grounding context as the change in the model’s generation confidence, $\gamma$, with and without conditioning on this context.
\begin{equation}
\small
\text{\grogu}_\theta(q,\mathbf{D_r}) = \gamma(y_g \mid q, \mathbf{D_r}) - \gamma(y_u \mid q) 
\label{eq:gu_overall}
\end{equation}
where $y_u$ and $y_g$ are the sequences generated when not conditioning/conditioning on the grounding documents, respectively.

\paragraph{Confidence Formulations:} We explore several formulations for the confidence measure. The most straightforward confidence measure can be derived by taking the negative values of the average perplexity and average entropy of the generated tokens.

\paragraph{Key-token Confidence:}
When an LLM generates an answer, there may be many tokens that are actually high-confidence in almost all the cases. For example, scaffolding phrases (e.g., ``My answer is ...''), parts copied from the question), stop words or sub-words that always follow other sub-words. These tokens may skew the confidence scores and affect the estimation of \grogu (\autoref{fig:non_answer_tokens} in \aautoref{append:examples} presents an example).

We propose using the confidence of only a subset of the generated tokens, which we define as \textit{key} tokens. We use entropy as an example and assume $\mathbf{D_r}$ only contains a single document ($D_r$) to simplify the notation and language.\footnote{The definitions based on perplexity and multiple documents are easy extensions.} We first define entropy of the key tokens (KeyEntropy), and then compute \grogu as the difference of the generated answer’s KeyEntropy with (and without) conditioning on the grounding document $D_r$. 

We start by computing the KeyEntropy for the grounded generation, i.e., $\text{KeyEntropy}(y_g \mid q, \mathbf{D_r})$, which corresponds to the first term in \autoref{eq:gu_overall}. Specifically, we obtain the generated answer $y_g$ under greedy decoding when conditioning on $\mathbf{D_r}$ and $q$. We then compute the token distributions for the same $y_g$ when conditioning only on $q$.\footnote{Note that we don't generate the ungrounded answer but we use the tokens from $y_g$.} A token $y_i$ in a generated sequence $y_g$ is a key token if

{\small\[
\left|H_\theta\left(y_i \mid q, D_r, y_{0, \ldots, y-1}\right)-H_\theta\left(y_i \mid q, y_{0, \ldots, y-1}\right)\right|>\alpha
\]}

\noindent where $\alpha$ is a chosen threshold that can be set empirically on a validation set, $H$ is the token-level entropy.\footnote{We use the standard definition. For the $i$-th token, $H=-\sum_{j \in V} p_j \text{log}(p_j)$, where $p_j$ is the probability for token $j$ in the vocabulary to be the $i$-th token, conditioned on the previous tokens in the sequence.} The intuition is that tokens that are not related to the core of the answer tend to have less change in their entropy, as for example, the tokens copied from the query. Thus, we can identify them by comparing the change in likelihood with and without the grounding context. We then only calculate the confidence scores on the key tokens. If no key tokens are identified (i.e., no changes exceed the threshold), we take the top K\% tokens with the highest entropy (instead of using entropy changes), where K is a hyper-parameter. The reason behind this is that when all tokens have small entropy changes, we find that using the token entropy directly gives slightly more robust performance than the entropy changes. Finally, KeyEntropy is the average entropy of the key tokens. \autoref{fig:keytoken_selection} in \aautoref{append:examples} shows an example.

It is worth noting that this definition of KeyEntropy does not directly apply to the ungrounded generation. Since we are interested in comparing different grounding contexts' utility given a specific query, the actual confidence of the ungrounded generation will be a common factor that can be discarded in the comparison. We provide one possible definition of ungrounded KeyEntropy in \aautoref{append:additional_discussion}, though it will not affect our experiments.
\section{Properties of Grounding Utility}
\label{sec:inference_only_tests}

We study the properties of \grogu with three tests, evaluating whether it can distinguish ground-truth documents from documents that do not contain an answer (\autoref{sec:golddoctest}) and whether the metric can capture relationships ignored by relevance scores (\autoref{sec:randdoctest} and \autoref{sec:samedoctest}). We follow \citet{cuconasuPowerNoiseRedefining2024} and use the test split of Natural Questions~\citep{kwiatkowski-etal-2019-natural} as our data source (unless otherwise specified) and Contriever~\citep{izacard2022unsuperviseddenseinformationretrieval} as our retriever for all three tests, but only the second test (\autoref{sec:randdoctest}) is inspired by \citet{cuconasuPowerNoiseRedefining2024}.

\subsection{\golddoctest}
\label{sec:golddoctest}

\begin{table}
\centering
\small
\renewcommand{\arraystretch}{0.1}
\
\begin{tabular}{b{0.5cm}b{1.3cm}b{0.9cm}b{0.9cm}b{0.9cm}b{0.5cm}}

\toprule
 &  & \multicolumn{2}{c}{\textbf{NQ}} & \multicolumn{2}{c}{\textbf{SQuAD}} \\
\cmidrule(lr){3-4} \cmidrule(lr){5-6}
\textbf{Model} & \textbf{Metric} & Gold/
Dist& Gold/
Rand& Gold/
Dist& Gold/
Rand\\
\midrule
\multirow{4}{*}{\textbf{Phi}} 
 & PPL        &\quad\quad71.3  &\quad\quad72.7  &\quad\quad87.1  &\quad\quad89.3 \\
 
 & KeyPPL     &\quad\quad76.9  &\quad\quad83.6  &\quad\quad 89.7  &\quad\quad 94.5 \\
 
 & Entropy&\quad\quad \cellcolor{green!15}74.7  &\quad\quad \cellcolor{green!15}76.2  &\quad\quad \cellcolor{green!15}89.2  &\quad\quad \cellcolor{green!15}91.5 \\
 
 & KeyEntropy&\quad\quad \cellcolor{green!15}77.9\textsuperscript{**} &\quad\quad \cellcolor{green!15}83.2\textsuperscript{**} &\quad\quad \cellcolor{green!15}90.7\textsuperscript{**} & \quad\quad\cellcolor{green!15}95.8\textsuperscript{**} \\
 
\midrule
\multirow{4}{*}{\textbf{Qwen}} 
 & PPL        &\quad\quad 79.8  &\quad\quad 85.2  &\quad\quad 82.5  &\quad\quad 90.6 \\
 & KeyPPL     &\quad\quad 80.2  &\quad\quad 87.1  &\quad\quad 83.2  &\quad\quad 91.2 \\
 & Entropy&\quad\quad \cellcolor{green!15}81.6  &\quad\quad \cellcolor{green!15}87.2  &\quad\quad \cellcolor{green!15}84.5  &\quad\quad \cellcolor{green!15}91.0 \\
 & KeyEntropy&\quad\quad \cellcolor{green!15}82.4\textsuperscript{*} &\quad\quad \cellcolor{green!15}88.8\textsuperscript{**} &\quad\quad \cellcolor{green!15}84.3  &\quad\quad \cellcolor{green!15}91.0 \\
\bottomrule
\end{tabular}
\caption{Win rates of gold vs. distractor/random with different metrics on Natural Questions (NQ) and SQuAD~\citep{rajpurkar2016squad100000questionsmachine}. We report statistical significance between Entropy and KeyEntropy (* and ** indicate p<0.1 and p<0.05 significance, respectively).}
\label{table:gtwinrate}
\end{table}

While \grogu is designed to capture utility beyond relevance for a specific LLM, we still want it to reflect some of the relationships based on ground-truth annotations. In order to measure this aspect, we create specific tests by leveraging existing datasets having annotations of the gold document that answers a query. In particular, we want to test whether \grogu assigns a higher utility to a ground-truth document ($D_{gt}$) that contains the answer and a lower utility to other documents. For each query, we curate a set of ``distractor'' ($D_{dst}$) and ``random'' ($D_{rnd}$) documents. $D_{dst}$ is the document with the highest retriever-assigned relevance but that does not contain the answer. For each query $q$, we compute the utilities of $D_{gt}$, $D_{dst}$, and $D_{rnd}$ under different \grogu formulations: i) PPL and KeyPPL, which use the perplexity of the tokens and key-tokens, respectively; ii) Entropy and KeyEntropy, which use the entropy and key-token entropy, respectively.
We report the win-rates of the gold document according to each \grogu formulation, e.g., if $\text{\grogu}_{\theta}(q, D_{gt})>\text{\grogu}_{\theta}(q, D_{dst})$, the gold document wins over the ``distractor.''
An effective metric should lead to high win-rates for the gold document.

\paragraph{Result} 
\autoref{table:gtwinrate} reports the win rates of the gold document against the distractor and the random document when tested with two LLMs: Phi-4 (14B) and Qwen-2.5-7B-Instruct. We refer to them as Phi and Qwen throughout the paper unless otherwise noted. We present additional experiments on two larger models in \aautoref{append:large_model_results}, which show similar trends.

We observe that the entropy-based metrics overall assign higher scores to gold passages than their perplexity-based counterparts. Key-token selection substantially increases the performance of both perplexity and entropy, and KeyEntropy overall leads to higher win rates. We use the sign test to measure whether the KeyEntropy method meaningfully improves over the simpler Entropy approach (rows corresponding to these methods are highlighted in green in \autoref{table:gtwinrate}). We find that the key-token selection variant often leads to statistically significant improvements in 6 of 8 comparisons (in the other two comparisons the two methods are not significantly different).

\subsection{\randdoctest}
\label{sec:randdoctest}

In order to further understand how relevance and utility differ in RAG, we use the same setting as in \citet{cuconasuPowerNoiseRedefining2024}'s experiments with recent LLMs. In particular, we test the setting of using a mixture of one gold document with four random/retrieved documents and measure how \grogu metrics capture the utility. Similarly to \citet{cuconasuPowerNoiseRedefining2024}, we find that contexts with 1 gold document and 4 random documents (\wrand) achieves a higher accuracy\footnote{We follow \citet{cuconasuPowerNoiseRedefining2024}'s steps to compute accuracy, which is a binary label.} than a mixture of 1 gold document and the top 4 documents returned by a retriever (\wretrieve). Although these two grounding contexts often give the same accuracy (`ties' in \autoref{table:pon}), there are many queries where conditioning on the \wretrieve context lead to an incorrect answer while conditioning on the \wrand context lead to a correct one (the `w/ rand wins' in the table). This suggests that a pure relevance-based metric would fail to reflect the utility of the documents.

\begin{table}[h!]
\centering
\small
\begin{tabularx}{\columnwidth}{l *{2}{>{\centering\arraybackslash}X}}
\toprule
 & Phi & Qwen \\
\hline
w/ retr Acc (\%) & 87.0 & 80.3 \\
w/ rand Acc (\%) & \textbf{91.4} & \textbf{88.2} \\
\hline
w/ retr wins & 58 & 93 \\
w/ rand wins & 186 & 321 \\
ties          & 2645 & 2475 \\
\bottomrule
\end{tabularx}
\caption{LLM accuracy (\%) when conditioned on \wretrieve and \wrand contexts and the counts of queries where these contexts lead to different accuracies.}
\label{table:pon}
\end{table}

A good grounding utility metric should predict this counter-intuitive result. To evaluate our \grogu, we extract test instances from \citet{cuconasuPowerNoiseRedefining2024}
where either \wrand generates a correct answer and \wretrieve generates an incorrect answer (or vice versa). For a query, we define the utility scores are \emph{concordant} if the grounding context (\wrand or \wretrieve) that generates the correct answer has higher utility. Similarly, \emph{discordant} is defined as the grounding context that generates the incorrect answer has higher utility.

Let $C$ and $D$ be the counts of concordant and discordant cases respectively. We apply the formula of Kendall's tau correlation $\tau=(C-D)/(C+D)$. This gives a rank correlation between -1 and 1. Intuitively, this reflects how much the observed impact of random documents agrees with the ``less than'' or ``greater than'' relationship of the confidence values.

As an additional measure of the relationship between \grogu and correctness, we predict which of two grounding contexts will generate the correct answer by comparing the \grogu scores. We report the binary classification performance of this classifier. Intuitively, the higher the classification performance, the better the grounding utility metric reflects the impact of the documents.

\paragraph{Result} As shown in \autoref{table:correl_w_acc}, all utility metrics have a positive correlation with the generation performance. KeyEntropy outperforms all the other formulations, achieving a correlation ($\tau$) of 0.320 for Phi and 0.377 for Qwen. This suggests that while there exist other factors not reflected in the utility scores that contribute to generation accuracy, KeyEntropy can indeed explain part of the performance differences between \wrand and \wretrieve. On the other hand, the relevance-based metric (mean relevance score of grounding documents) has a negative correlation by always assigning a lower score to the \wrand context, completely failing to reveal its positive impact.
\begin{table}[h]
\centering
\small
\begin{tabularx}{\columnwidth}{l l c c c}
\hline
\textbf{Model} & \textbf{Metric} & \textbf{$\tau$} & \textbf{Acc (\%)} & \textbf{F1 (\%)} \\
\hline
\multirow{5}{*}{\textbf{Phi}} 
  & PPL        & 0.172  & 58.6 & 52.3 \\
  & KeyPPL     & 0.238  & 61.9 & 57.7 \\
  & Entropy    & 0.295  & 64.8 & 56.6 \\
  & KeyEntropy & \textbf{0.320} & \textbf{66.0} & \textbf{59.6} \\
  & Relevance  & -0.525 & 23.8 & -- \\
\hline
\multirow{5}{*}{\textbf{Qwen}} 
  & PPL        & 0.266  & 63.3 & 55.6 \\
  & KeyPPL     & 0.275  & 63.8 & 56.8 \\
  & Entropy    & 0.348  & 67.4 & 58.1 \\
  & KeyEntropy & \textbf{0.377} & \textbf{68.8} & \textbf{60.6} \\
  & Relevance  & -0.551 & 22.5 & -- \\
\hline
\end{tabularx}
\caption{Simple classifier performance and the correlation between utility and accuracy ($\tau$). Relevanced-based classifier would always predict the same label thus it's F1 is undefined.}
\label{table:correl_w_acc}
\end{table}

\subsection{\samedoctest}
\label{sec:samedoctest}
\begin{figure}[h]
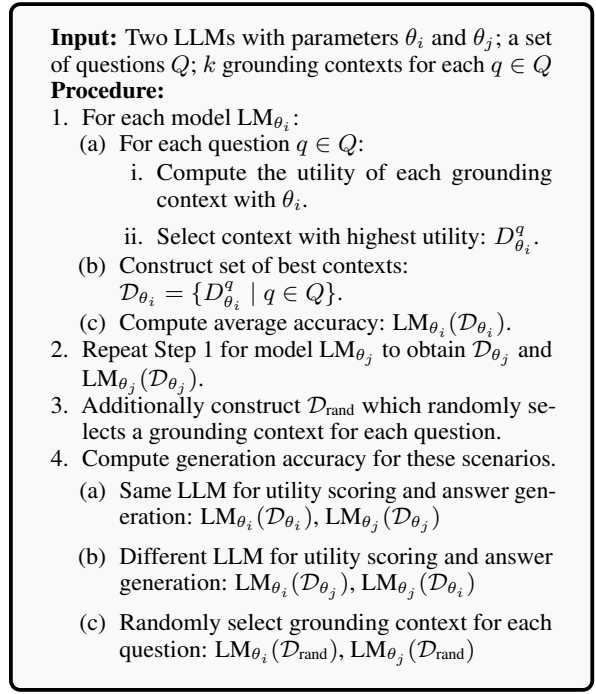

\centering
\begin{minipage}{\columnwidth}
\small
\begin{tcolorbox}[colback=gray!5,colframe=black]
\textbf{Input:}  Two LLMs with parameters $\theta_i$ and $\theta_j$; a set of questions $Q$; $k$ grounding contexts for each $q\in Q$ \\
\textbf{Procedure:}  
\begin{enumerate}[leftmargin=*, nolistsep]
    \item For each model $\text{LM}_{\theta_i}$:  
    \begin{enumerate}[leftmargin=14pt, noitemsep, nolistsep]
        \item For each question $q \in Q$:  
        \begin{enumerate}[leftmargin=14pt, itemsep=4pt]
            \item Compute the utility of each grounding context with $\theta_i$.  
            \item Select context with highest utility: $D_{\theta_i}^q$.  
        \end{enumerate}
        \item Construct set of best contexts: \\$\mathcal{D}_{\theta_i} = \{D_{\theta_i}^q \mid q \in Q\}$.  
        \item Compute average accuracy: $\text{LM}_{\theta_i}(\mathcal{D}_{\theta_i})$.  
    \end{enumerate}
    \item Repeat Step 1 for model $\text{LM}_{\theta_j}$ to obtain $\mathcal{D}_{\theta_j}$ and $\text{LM}_{\theta_j}(\mathcal{D}_{\theta_j})$.  
    \item Additionally construct $\mathcal{D}_{\text{rand}}$ which randomly selects a grounding context for each question.
    \item Compute generation accuracy for these scenarios. 
    \begin{enumerate}[leftmargin=14pt, itemsep=4pt]
        \item Same LLM for utility scoring and answer generation: $\text{LM}_{\theta_i}(\mathcal{D}_{\theta_i})$, $\text{LM}_{\theta_j}(\mathcal{D}_{\theta_j})$ 
        \item Different LLM for utility scoring and answer generation: $\text{LM}_{\theta_i}(\mathcal{D}_{\theta_j})$, $\text{LM}_{\theta_j}(\mathcal{D}_{\theta_i})$ 
        \item Randomly select grounding context for each question: $\text{LM}_{\theta_i}(\mathcal{D}_{\text{rand}})$, $\text{LM}_{\theta_j}(\mathcal{D}_{\text{rand}})$ 
    \end{enumerate}
\end{enumerate}
\end{tcolorbox}
\end{minipage}

\caption{Procedure to test the utility of the same document for different LLMs.}
\label{fig:same_doc_algo}
\end{figure}

Different LLMs may derive different levels of utility from the same context, i.e., a document that helps one model may not be useful for another one. In this experiment, we construct synthetic contexts that contain the same information but that vary in the physical layout of that information. We accomplish this by first selecting ten documents for each question (including the gold document and nine retrieved documents) and then creating three alternative contexts by varying where the gold document sits among the documents (first, fifth, or last). Our objective is to demonstrate that our \grogu scores can select contexts that improve answer correctness over a random choice and that the preferences indicated by \grogu scores for one model do not apply to other models. \autoref{fig:same_doc_algo} presents the test procedure.

\begin{table}[h!]
\centering
\small
\begin{tabular}{lcc|c}
    \toprule
    & \multicolumn{3}{c}{\textbf{Metric}} \\
    \midrule
   & Entropy & KeyEntropy & Random\\
    ${\text{Phi}}(\mathcal{D}_\text{Phi})$       & \textbf{73.3} & \textbf{74.7} & \multirow{2}{*}{72.9} \\
    ${\text{Phi}}(\mathcal{D}_\text{Qwen})$      & 72.1 & 72.5 \\   
   
    \midrule
    ${\text{Qwen}}(\mathcal{D}_\text{Qwen})$      & \textbf{68.8} & \textbf{69.3} & \multirow{2}{*}{66.0}\\
    ${\text{Qwen}}(\mathcal{D}_\text{Phi})$       & 66.1 & 66.3 &   \\

    
    \bottomrule
\end{tabular}
\caption{Generation Accuracy for different contexts scenario. We present the random baselines in the last column, which correspond to $\text{Phi}(\mathcal{D}_{\text{rand}})$ and $\text{Qwen}(\mathcal{D}_{\text{rand}})$.}
\label{table:same_doc_diff_utils}
\end{table}

\paragraph{Result:} \autoref{table:same_doc_diff_utils} presents the result of this test. We focus on grounding utility based on Entropy and KeyEntropy.\footnote{Other metrics achieve similar results (\aautoref{append:samedoc_test_additional}).} $\text{Phi}(\mathcal{D}_{\text{Phi}})$ denotes the responses generated using Phi when given contexts selected by Phi's \grogu scores. Similarly, $\text{Phi}(\mathcal{D}_{\text{Qwen}})$ denotes responses generated by Phi given contexts selected by Qwen's \grogu scores. (We use similar notation for the third and fourth rows for Qwen). We highlight two conclusions from these results. First, using our \grogu scores to select which context to use for response generation improves over a random choice (this aligns with findings in the previous two experiments). Second, for a given model, using the best context for another model does not perform as well as using the best context selected by this model's own \grogu score (all these accuracy gaps are statistically significant).

This experiment establishes that LLMs can prefer different physical layouts of the same information (in this case different gold passage positions) and that \grogu can help to select the layout that leads to higher accuracy. We expect that other variations in how information is expressed (e.g. layout, markup languages, style, and so on) will also lead to different impacts on different models. 

\section{Query-Rewriter Training}
\label{sec:training}

In \autoref{sec:inference_only_tests}, we observed that KeyEntropy achieves the best overall performance across the tests, so we use it as our \grogu metric to train a query-rewriter.
Under this scenario, a user and a RAG system have a multi-turn conversation. The user's questions will be based on past turns and the system needs to rewrite the query to add missing information before sending it to the retriever. Query-rewriting is an important step for RAG since human interaction with modern LLMs are highly conversational. \autoref{fig:query-rewriting}, \aautoref{append:illustration} provides an illustration.
In this section, we present our method of training a query-rewriter by using \grogu  without any annotated labels.
\subsection{Training Data Creation}
Our data preparation and training resemble \citet{yoonAskOptimalQuestions2025}’s supervised pipeline (RetPO) but we replace their steps that require gold annotations with \grogu. In the original method, RetPO prompts GPT-4 with three different strategies to achieve a diverse sample of 25 rewrites for each question. It then sends those rewrites to a retriever and measure the quality of the rewrite based on the rank of the gold passage in the top 100 passages the rewrite retrieves. Informed by this retrieval performance, it selects the best rewrites to construct a dataset for SFT warmup and a preference dataset for DPO training (by pairing better and worse rewrites based on their gold passage ranks).

Unlike RetPO, we do not assume access to gold passage labels; instead, we use \grogu to evaluate retrieval success. For each rewrite, we take the top 10 retrieved documents and send them to an LLM generator from which we derive a utility score. We take the rewrite corresponding to the highest \grogu for SFT warmup and we pair the rewrites with the highest and lowest \grogu  for DPO. Because we want the preferred and dispreferred rewrites in our DPO dataset to be different enough, we filter out 50\% preference pairs based on the \grogu gap between the preferred and dispreferred (removing those with smaller gaps). The rest of the procedure and the raw data follow the RetPO experiments. We discuss these implementation details further in \aautoref{append:implementation}.

\subsection{Training Experiment}

\begin{table*}[h]
    \footnotesize
    \centering
    \begin{threeparttable}
    \renewcommand{\arraystretch}{0.6}

    \begin{tabular*}{0.93\textwidth}{cl|cccc|cccc}
        \toprule
        & &   \multicolumn{4}{c}{\textbf{TopiOCQA}} &  \multicolumn{4}{c}{\textbf{QReCC}}  \\ 
        \textbf{Type} & \textbf{Method} & \textbf{MRR} & \textbf{R@10}  & \textbf{R@20} & \textbf{R@100} & \textbf{MRR} & \textbf{R@10} & \textbf{R@20} & \textbf{R@100} \\
        \midrule 
        \multirow{9}{*}{\rotatebox[origin=c]{90}{\textbf{Sparse (BM25)}}} 
        
        & \textsc{Qwen - \texttt{Original}}   &14.6 & 27.8 & 35.5 & 52.7 & 27.7 & 43.6 & 50.7 & 65.4 \\
        & \textsc{\qquad\quad~~\texttt{Base SFT}} &17.4 & 32.8 & 41.6 & 60.5  &32.4 & 51.4 & 60.1 & 75.5   \\
        & \textsc{\qquad\quad~~\texttt{\grogu}} &25.7 & 46.1 & 54.7 & 72.0 &\textbf{45.9} & \textbf{67.2} & \textbf{74.9} & \textbf{85.6} \\
        & \textsc{Phi - \texttt{Original}}   &16.6  &31.3  &39.1  &58.0  &36.8	&53.7	&61.9	&75.3 \\
        & \textsc{\qquad\quad~~\texttt{Base SFT}} &15.7 & 29.0 & 37.8 & 55.1 & 32.1 & 50.4 & 58.6 & 73.5\\
        & \textsc{\qquad\quad~~\texttt{\grogu}}     &\textbf{27.2} & \textbf{48.8} & \textbf{58.6} &\textbf{74.8} & 45.8 & 66.2 & 73.7 & 84.3\\
        \cmidrule(lr){2-10}
        & RetPO   & \textbf{28.3} & 48.3 & -  & \textbf{73.1} & 50.0 & 69.5 & - & 89.5 \\
        & ConvGQR & 12.4 & 23.8 & -    & 45.6 & 44.1 & 64.4 & - & 88.0 \\
        & IterCQR & 16.5 & 29.3 & -    & 54.1 & 46.7 & 64.4 & - & 85.5 \\
        & ADACQR  & \textbf{28.3} & \textbf{48.9} & -    & 71.2 & \textbf{55.1} & \textbf{76.5} & -    & \textbf{93.7}\\
        \midrule 
        \multirow{9}{*}{\rotatebox[origin=c]{90}{\textbf{Dense (ANCE)}}} 
        & \textsc{Qwen - \texttt{Original}} &28.1 & 47.7 & 54.1 & 65.1 & 29.6 & 45.7 & 51.0 & 59.9\\
        & \textsc{\qquad\quad~~\texttt{Base SFT}} &33.9 & 56.8 & 64.2 & 76.1  &35.2 & 54.1 & 61.1 & 71.2\\
        & \textsc{\qquad\quad~~\texttt{\grogu}} &37.8 & \textbf{61.7} & \textbf{68.7} & \textbf{79.8} &\textbf{40.2} & \textbf{60.8} & \textbf{67.5} & \textbf{76.8}\\
        & \textsc{Phi - \texttt{Original}}   &31.6	&53.6	&60.6	&73.1  &36.0	&54.9	&61.3	&70.6	\\
        & \textsc{\qquad\quad~~\texttt{Base SFT}} &29.7 & 51.3 & 59.6 & 72.6 & 32.9 & 51.8 & 58.5 & 69.4 \\
        & \textsc{\qquad\quad~~\texttt{\grogu}}     &\textbf{38.1} & 61.1 & 68.3 & 79.6 & 39.4 & 59.8 & 67.1 & 76.3\\
        \cmidrule(lr){2-10} 
        & RetPO   & 30.0 & 49.6 & - & 68.7 & 44.0 & 66.7 & - & \textbf{84.6} \\ 
        & ConvGQR & 25.6 & 41.8 & - & 58.8 & 42.0 & 63.5 & - & 81.8 \\
        & IterCQR & 26.3 & 42.6 & - & 62.0 & 42.9 & 65.5 & - & 84.1 \\
        & ADACQR  & \textbf{38.5} & \textbf{58.4} & - & \textbf{75.0} & \textbf{45.8} & \textbf{67.3} & - & 83.8 \\
        \bottomrule 
    \end{tabular*}
    \end{threeparttable}
    \caption{Sparse and dense retriever performance on TopiOCQA and QReCC. Baseline models' performance scores are those reported in their original papers, which do not report Recall@20.}
    \vspace{-2mm}
    \label{table:main}
\end{table*}

We use Phi-4 (14B) and Qwen-2.5-7B-Instruct for our main experiments and additionally show our method generalizes to larger models in \aautoref{append:large_model_results}. We choose these models because they are the mainstream models with completely open licenses. We further discuss our model choices in \aautoref{append:model_choice}. The ``Phi'' and ``Qwen'' in the following section refer to the 7B Qwen model and the 14B Phi model. 

The most comparable baseline RetPO uses llama2-7b-chat. Thus we attempted to use Qwen and Phi to reproduce their experiments. We reproduced half of their experiments (those involving BM25) because they only released the curated preference data for these experiments. We observed similar performance with Qwen and Phi, not outperforming the results reported in their paper. Because of this observation and not having the curated preference data to reproduce their ANCE experiments, we use the performance reported in the original RetPO paper as our main baseline.

We exactly follow \citet{yoonAskOptimalQuestions2025} for our evaluation. We test our method on the sparse retriever BM25 and the dense retriever ANCE~\citep{xiong2020approximatenearestneighbornegative}. We then benchmark the RAG system on top of: \topiocqa~\citep{adlakhaTopiOCQAOpendomainConversational2022a} and \qrecc~\citep{ananthaOpenDomainQuestionAnswering2021}. More details about the setup are provided in \aautoref{append:implementation}.

\paragraph{Baselines} We compare our method with representative baselines that use annotated labels.
The most comparable baseline is RetPO as we closely follow their pipeline. Other baselines we report include: ConvGQR~\citep{moConvGQRGenerativeQuery2023}, which uses gold answers and combines query rewriting with expansion; IterCQR~\citep{jangIterCQRIterativeConversational2024}, which uses gold passage annotations and iterative training; and ADACQR~\citep{laiAdaCQREnhancingQuery2025}, which uses gold passage annotations and aligns the rewriter with both the sparse and dense retrievers.

These methods cover the common baselines used in studies of conversational query-rewriting. We include a discussion of other related but not comparable methods and their performance in \aautoref{append:not_comparable_methods}. These methods are Query2Doc~\citep{wang-etal-2023-query2doc}, Promptriever~\citep{weller2024promptriever}, and ConvSearch-R1~\citep{zhuConvSearchR1EnhancingQuery2025}. Query2Doc and Promptriever are not designed for conversational query-rewriting, and thus performed worse than our methods and several of the common baselines. ConvSearch-R1 is a recent reasoning-based query-rewriting method that requires gold labels. It uses online RL and reasoning chains, thus requiring substantially more compute for both training and inference than our method and other baselines. We provide further discussion in \aautoref{append:not_comparable_methods}.

\paragraph{Evaluation} We use retrieval metrics based on gold passages (MRR and Recall@K) for retrieval results. We evaluate the LLM's answers with a GPT-4o-mini judge that compares the generated answers with gold answers based on the judge prompt from \citet{ni2024mixevalderivingwisdomcrowd}. We also report word overlap metrics for LLM's answers, like F1 and Exact Match, as in \citet{adlakhaTopiOCQAOpendomainConversational2022a}.
\subsection{Training Results}
We report the retrieval and generation performance. In the provided tables, \texttt{ORIGINAL} refers to the untrained original models, \texttt{BASE SFT} refers to directly distilling from randomly sampled GPT4o rewrites from the same raw dataset as our method, instead of doing data selection with \texttt{\grogu}. Finally,  \texttt{\grogu} refers to our method described above, i.e., an SFT warmup followed by a DPO stage. \autoref{append:implementation} provides details about our training setup.

\paragraph{Retrieval Results} \autoref{table:main} reports the retrieval performance of different methods. Overall, we see consistent improvements on all the retrieval metrics with our method compared with not using utility-guided data selection (\texttt{Base SFT}). This trend is consistent across retrievers, LLMs, and benchmarks. For example, under the BM25+\qrecc setting, Qwen (\texttt{\grogu}) outperforms the \texttt{Base SFT} baseline by 13.5 in MRR and Phi (\texttt{\grogu}) also outperforms it by 13.7 in MRR. Compared with \texttt{Original}, \texttt{\grogu} also improves by up to 18.2 (Qwen, BM25+\qrecc).
Importantly, our method achieves performance close to or sometimes better than some reference-based baselines. For example, for the BM25+\topiocqa setting, Qwen (\texttt{\grogu}) and Phi (\texttt{\grogu}) have performances close to the most comparable baseline (RetPO) in all metrics. \texttt{\grogu} also outperforms RetPO under the ANCE+\topiocqa setting. Overall, the results suggest that there is a need to finetune a query rewriter (given our big improvement over \texttt{ORIGINAL}), and our method offers an easy, annotation-free way for such training.

\paragraph{Generation Performance}
Generation performance has not been an informative metric for conversational query-writing benchmarks, due to the free-form, open-ended nature of the questions. The related works, including the recently published ones~\citep{yoonAskOptimalQuestions2025, laiAdaCQREnhancingQuery2025, zhuConvSearchR1EnhancingQuery2025}, all only use retrieval performance to evaluate their methods. In \aautoref{append:reader_performance}, we provide the generation of performance of our trained models just for the sake of completeness, while noting that the previously presented retrieval performance is a more reliable measure. Overall, we observe that our method is able to consistently provide statistically significant improvements in the downstream LLM's generation accuracy, by up to 9.4 percentage points. The magnitude of improvement is smaller than those with the retrieval metrics, likely a result of the accuracy metric underestimating the actual performance in free-form QA. Yet, they still show that our annotation-free method can provide sizable improvements in generation accuracy. We discuss the concept and results of generation performance in greater detail in \aautoref{append:reader_performance}.

\section{Related Work}
\label{sec:related_and_future_work}
Prior methods to improve RAG systems have primarily used metrics based on annotated labels. Some use the rank of the gold passages to quantify retrieval performance~\cite[e.g.,][]{jangIterCQRIterativeConversational2024,laiAdaCQREnhancingQuery2025}. These metrics have served as reward functions to build DPO training data~\citep{yoonAskOptimalQuestions2025} or for online RL~\citep{zhuConvSearchR1EnhancingQuery2025}. Other works send documents to the downstream LLM and compare the generated answer with the gold one(s), using metrics such as accuracy, Exact Match, and the likelihood of gold answer~\citep{moConvGQRGenerativeQuery2023, zhangAdaptiveQueryRewriting2024, sunDynamicRAGLeveragingOutputs2025}. These works also improve an individual component of RAG, e.g., the retriever~\citep{salemiSearchEngineMachines2024}) or the reranker~\citep{suDRAGINDynamicRetrieval2024}. In this paper, we focus on applying an annotation-free metric to improve the query-rewriter; we expect it to be applicable to these other components too. Finally, our metric is based on the idea of key token extraction, which is also used in \citet{fangWhatWrongPerplexity2025} but they focus on evaluating language modeling performance in the long-context setting by using the perplexity assigned to reference sequences. We instead apply key token selection to measure generation confidence and design it for the RAG setting.

\section{Conclusion}
\label{sec:conclusion}
In this work, we define \grogu, a metric for the utility of documents used as context of an LLM in RAG settings. \grogu is defined by automatically selecting key tokens in the grounded generations and computing the difference in entropy between the grounded and ungrounded generation. We show that \grogu captures nuances that relevance scores ignore. Using query-rewriting as an example task, we show that \grogu\ identifies effective preference data for DPO training. The data selected through \grogu\ can improve a query-rewriter by generating better queries to retrieve more useful documents, which allows for more accurate generation with the downstream LLM.

\section*{Limitations}
In this paper, we try to re-think the retrieval quality evaluation from the perspective of utility to an LLM in a RAG setting, assuming no access to any references/labels. To show this, we dedicate a substantial part of the paper to the motivation and the analysis of the properties of our utility metric \grogu with ad-hoc experiments. 
From a practical perspective, we then test the metric on a single task, i.e., query-rewriting, showing robust improvements in both retrieval and generation performance. We didn't specifically explore other tasks in this paper, which is already dense in content, but we expect \grogu to be also applicable to other components of RAG systems, e.g., the re-ranking stage.

As we pointed out throughout the paper, evaluating open-ended, free-form generation by comparing the generated answer with the gold one(s) can oftentimes underestimate the actual model performance. We still report this evaluation as part of our results to provide a complete picture but we dedicate more space to the retrieval performance as it is more suitable to show the impact of our method.

\section*{Ethics Statement}
Any Retrieval Augmented Generation system, regardless of retrieval quality, may still accidentally include misleading, non-factual, or other inappropriate documents that may be used by the downstream LLM to produce harmful content. Therefore, these systems should be deployed with caution and users should be informed of these limitations. Additionally, for some application scenarios, external modules may be necessary to detect inappropriate contents and prevent them from being propagated to the LLM's generation.

\bibliography{grounding_utils}

\appendix

\section{Additional Illustration}
\label{append:illustration}
We present figures of additional illustration in this appendix.
\begin{figure}[h]
\centering
\includegraphics[width=0.95\columnwidth]{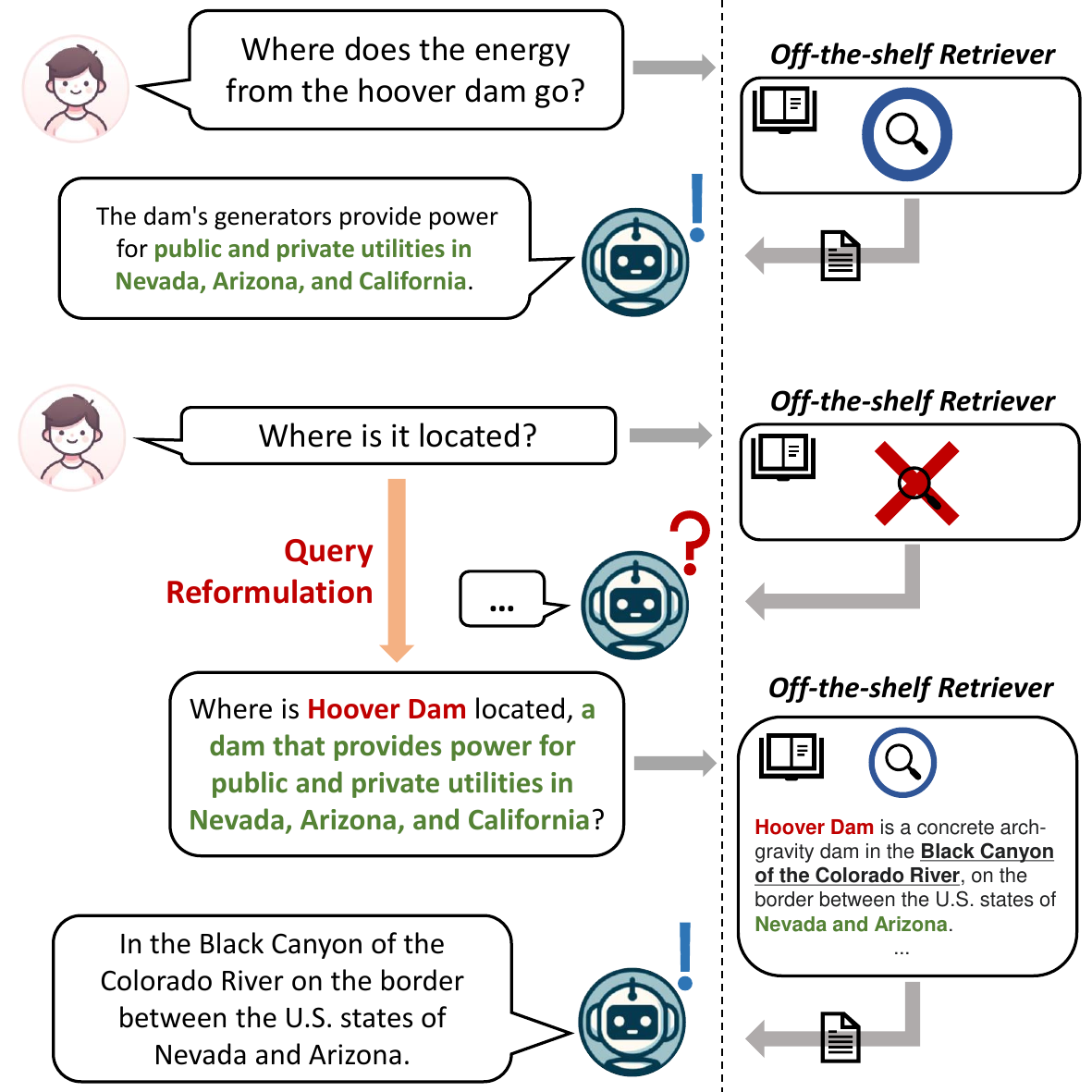}
\caption{Query-rewriting illustration from \citet{jangIterCQRIterativeConversational2024}. The user's queries are dependent on the conversation history and need to be reformulated into stand-alone queries for the retriever.}

\label{fig:query-rewriting}
\end{figure}

\section{Examples}
\label{append:examples}
\begin{figure*}[t]
\centering
\fbox{%
\parbox{0.95\textwidth}{%
\textbf{Question:} Who hosted They Think It’s All Over?\\
\textbf{Gold Answers:} [`Lee Mack’, `Nick Hancock’] \\

Grounded on a retrieved passage that \textbf{does not} contain the answer
\begin{itemize}
    \item Generated Answer: Jonathan Ross hosted “They Think It's All Over.” \textcolor{red}{\xmark}
    \item Entropy: 0.466
\end{itemize}

Grounded on a gold passage
\begin{itemize}
    \item Generated Answer: Nick Hancock, followed by Lee Mack  \textcolor{ForestGreen}{\cmark}
    \item Entropy: 0.569
\end{itemize}

\textbf{Entropy by token:} \\
\includegraphics[width=0.9\textwidth, trim=0 60 0 60, clip]{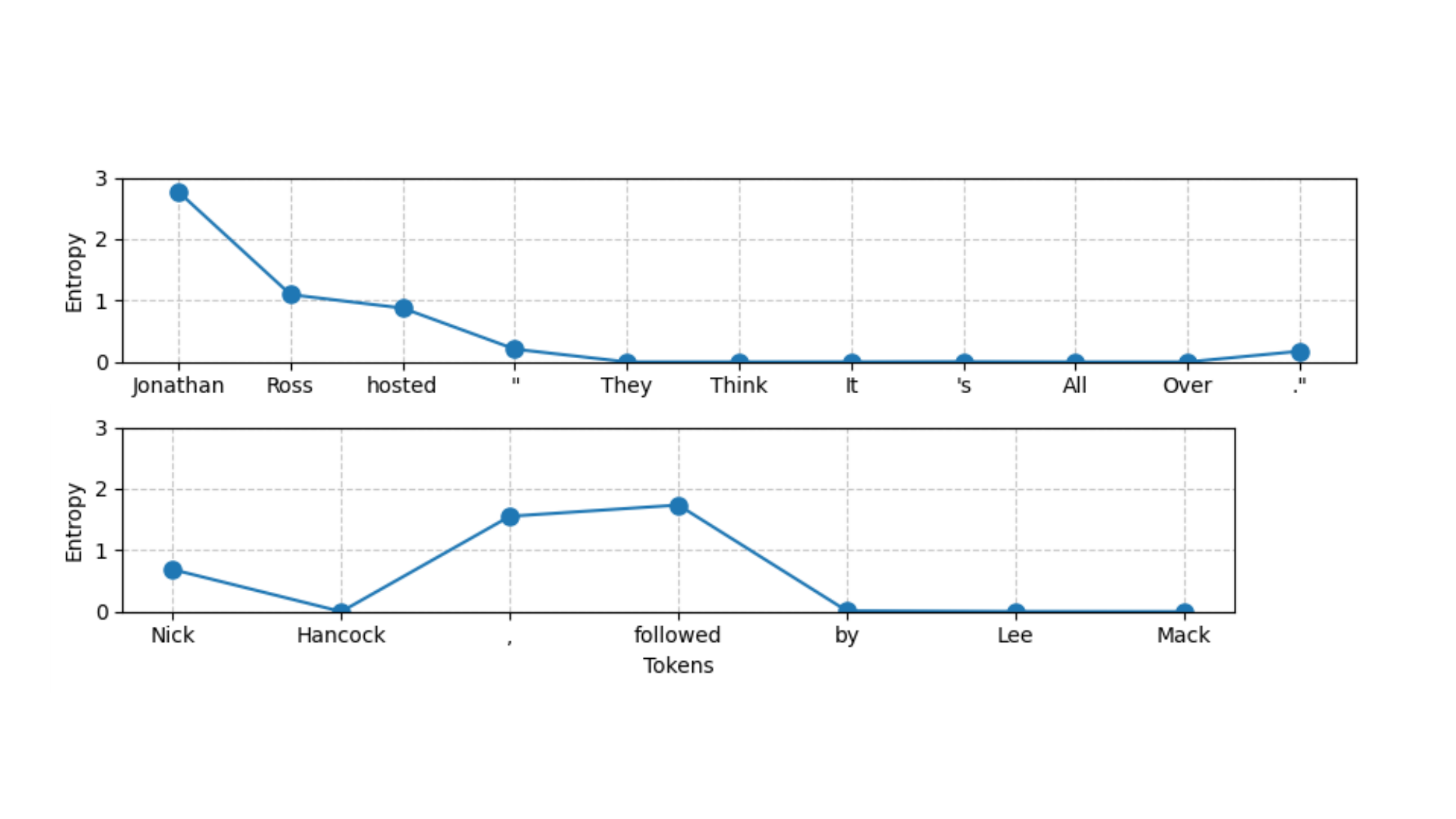}
}%
}
\caption{Example showing that non-answer tokens can skew the uncertainty/confidence measure. The model answers the question correctly when grounded on the gold passage but the average entropy is higher, indicating a lower confidence. This is because the incorrect answer repeats part of the question which skews down the average entropy of the generation.}
\label{fig:non_answer_tokens}
\end{figure*}

\begin{figure*}[t]
\centering
\fbox{%
\parbox{0.95\textwidth}{%
\textbf{Question:} Who hosted They Think It’s All Over?\\
\textbf{Gold Answers:} [`Lee Mack’, `Nick Hancock’] \\
{\small
Tokens selected by our KeyEntropy algorithm for calculation are in \textcolor{blue}{blue}.
}

Grounded on a retrieved passage that \textbf{does not} contain the answer
\begin{itemize}
    \item Generated Answer: \textcolor{blue}{Jonathan Ross hosted “}They Think It's All Over\textcolor{blue}{.”} \textcolor{red}{\xmark}
    \item Entropy: 0.466
    \item KeyEntropy: 1.024
\end{itemize}

Grounded on a gold passage
\begin{itemize}
    \item Generated Answer:  \textcolor{blue}{Nick Hancock, followed by Lee Mack} \textcolor{ForestGreen}{\cmark}
    \item Entropy: 0.569
    \item KeyEntropy: 0.569
\end{itemize}

\textbf{Entropy by token:} \\
\includegraphics[width=0.9\textwidth, trim=0 60 0 60, clip]{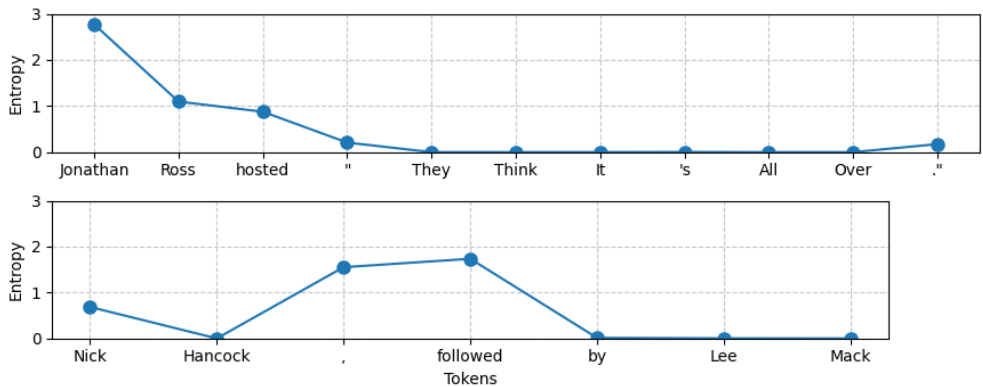}
}%
}
\caption{Example showing that KeyEntropy mitigates the issue of non-answer tokens. The model answers the question correctly when grounded on the gold passage but the average entropy is higher, indicating a lower confidence. KeyEntropy identifies and omits the tokens in the incorrect answer that repeat part of the question, which skews the entropy calculation.  }
\label{fig:keytoken_selection}
\end{figure*}

We present additional examples in \autoref{fig:non_answer_tokens} and \autoref{fig:keytoken_selection}.

\section{Additional Discussion}
\label{append:additional_discussion}
\subsection{Definition of ungrounded generation confidence} 
We define the key tokens of the ungrounded generation to satisfy
 \[
 \small
  H_\theta(y_i \mid q, D_r, y_{0, \ldots y-1})>\alpha,
 \]
where $y$ is the ungrounded generation. If no tokens satisfy this criterion, the top K\% tokens in entropy will be used. Under this definition, tokens with small entropy will not contribute to our uncertainty estimation. This helps mitigate the spurious correlations of non-key tokens because we empirically find that when the LLM is not given a grounding context, the low-entropy tokens in the generated answer are mostly non-answer tokens. We note that this simple definition should only be used for ungrounded generations because for a grounded generation, an important answer token may actually have low entropy due to the presence of the same/related tokens in the grounding document.

\subsection{Model Choices} 
\label{append:model_choice}
The models covered in this paper are the mainstream models that have completely open licenses (Apache and MIT licenses) and have model sizes that we can efficiently train with our compute resources. Some other “open models” were not chosen due to the legal constraints of their specific terms and conditions, which we cannot further comment on at this point due to the anonymity requirement. We will provide further explanation in our non-anonymous camera-ready version.

\section{Additional Results}
\label{sec:additional_results}
\subsection{\samedoctest}
\label{append:samedoc_test_additional}
\autoref{table:samedoc_test_ppl} presents additional results on the \samedoctest experiment introduced in \autoref{sec:samedoctest}. The same analysis with entropy-based metrics from \autoref{sec:samedoctest} still apply to the perplexity-based metrics. We see that the perplexity-based utility metrics can also help identifying the contexts that give higher performance than the random baseline and an LLM generator achieves the best performance if the utility scorer is the same model (e.g.,  ${\text{Phi}}(\mathcal{D}_\text{Phi})$  instead of  ${\text{Phi}}(\mathcal{D}_\text{Qwen})$). This further supports our analysis in \autoref{sec:samedoctest}. In general, using KeyPPL to select grounding contexts gives the higher performance than just standard perplexity, further indicating the benefits of key token selection. We also note that when Phi is used as both the generator and utility scorer, KeyPPL leads to slightly better performance than the corresponding KeyEntropy from \autoref{table:same_doc_diff_utils} (75.3 vs. 74.7) but this difference is not statistically significant and KeyEntropy still outperforms KeyPPL most of the time across the tasks.

\begin{table}[h!]
\centering
\small
\begin{tabular}{lcc|c}
    \toprule
    & \multicolumn{3}{c}{\textbf{Utility Metric}} \\
    \midrule
   & PPL & KeyPPL & Random\\
    ${\text{Phi}}(\mathcal{D}_\text{Phi})$       & \textbf{74.0} & \textbf{75.3} & \multirow{2}{*}{72.9} \\
    ${\text{Phi}}(\mathcal{D}_\text{Qwen})$      &  71.9&  72.4\\   
   
    \midrule
    ${\text{Qwen}}(\mathcal{D}_\text{Qwen})$      & \textbf{68.4} & \textbf{69.2} & \multirow{2}{*}{66.0}\\
    ${\text{Qwen}}(\mathcal{D}_\text{Phi})$       & 65.8&  66.5&  \\

    
    \bottomrule
\end{tabular}
\caption{Generation accuracy for different scenarios of the \samedoctest test for perplexity-based utility metrics}
\label{table:samedoc_test_ppl}
\end{table}

\subsection{Downstream Generator Performance}
\label{append:reader_performance}
We measure the performance of the downstream LLM that uses the documents retrieved by the rewrites. Because word-overlap-based metrics (Precision, Recall, F1 and Exact Match) are problematic for evaluating free-form generation~\citep{risch-etal-2021-semantic},\footnote{The evaluation of free-form, open-ended generation is difficult because they often have many correct answers~\citep{ananthaOpenDomainQuestionAnswering2021} while the benchmarks come with only one or few gold answers. Thus it can under-estimate the LLM's accuracy. LLM judge mitigates this by better recognizing semantically equivalent answers but does not solve the issue fully.} we focus on the accuracy judged by the LLM judge. We adapt the LLM-judge for QA tasks from \citet{ni2024mixevalderivingwisdomcrowd} to evaluate the generator’s answers. We use GPT-4o-mini as the judge and prompt it to produce a binary label for each of the generator’s answer. An answer is considered correct as long as the LLM judge matches any of the gold answers. The LLM judge’s output is reported as accuracies (“acc”) in \autoref{table:reader}. The LLM judge is more suitable for the free-form, open-ended setting than the word-overlap metrics, though it would still underestimate the generation performance when there are multiple semantically different correct answers but the provided gold answers are limited.

\definecolor{lightred}{RGB}{255,230,230}
\begin{table*}[t!]
    \centering
    \footnotesize
    \begin{threeparttable}
    \begin{tabular*}{0.93\textwidth}{cl|cccc>{\columncolor{lightred}}c|cccc>{\columncolor{lightred}}c}
        \toprule
        & &   \multicolumn{5}{c}{\textbf{TopiOCQA}} &  \multicolumn{5}{c}{\textbf{QReCC}}  \\ 
        \textbf{Type} & \textbf{Method.} & \textbf{EM}  & \textbf{f1\textsuperscript{*}} & \textbf{prec\textsuperscript{*}} & \textbf{recall\textsuperscript{*}}  & \textbf{Acc} & \textbf{EM}  & \textbf{f1\textsuperscript{*}} & \textbf{prec\textsuperscript{*}} & \textbf{recall\textsuperscript{*}} &\textbf{Acc}\\
        \midrule \midrule
        \multirow{9}{*}{\rotatebox[origin=c]{90}{\quad\quad\quad\textbf{BM25}}} 
        
        & \textsc{Qwen - \texttt{Original}}       & 17.7 & 39.8 & 47.5 & 46.3 & 48.8 & 1.0 & 28.8 & 23.8 & 49.3 & 51.4\\
        & \textsc{\qquad\quad~~\texttt{Base SFT}} & 19.0 & 42.5 & 50.3 & 48.7 & 52.4 & 1.1 & 30.8 & 25.7 & 51.6 & 56.0\\
        & \textsc{\qquad\quad~~\texttt{\grogu}}     & 20.3 & 44.7 & 52.6 & 51.3 & 56.2 & 1.2 & 32.3 & 26.8 & 53.9 & 60.5 \\
        & \textsc{Phi - \texttt{Original}}        &5.7	&32.3	&27.0	&63.6 &56.4 &0.4	&32.9	&25.7	&54.2 &59.2\\
        & \textsc{\qquad\quad~~\texttt{Base SFT}} &5.5 & 32.2 & 26.8 & 63.7 & 57.2 & 0.2 & 31.9 & 24.8 & 53.0 & 58.1\\
        & \textsc{\qquad\quad~~\texttt{\grogu}}     &6.6 & 35.9 & 30.0 & 69.8 & \textbf{65.8} & 0.4 & 34.1 & 26.5 & 56.5 & \textbf{62.6}\\
        \midrule \midrule
        \multirow{9}{*}{\rotatebox[origin=c]{90}{\quad\quad\quad\textbf{ANCE}}} 
        & \textsc{Qwen - \texttt{Original}} &19.2 & 43.1 & 50.5 & 50.5 & 54.3 & 0.9 & 27.9 & 22.9 & 48.7 & 50.4\\
        & \textsc{\qquad\quad~~\texttt{Base SFT}} & 20.9 & 46.7 & 54.0 & 55.1 & 60.2 & 1.1 & 30.1 & 24.7 & 51.6 & 55.7 \\
        & \textsc{\qquad\quad~~\texttt{\grogu}}    & 20.8 & 46.9 & 54.6 & 54.7 & 60.7 & 1.1 & 31.1 & 25.6 & 52.9 & 58.3\\
        & \textsc{Phi - \texttt{Original}}       &6.2	&36.0	&29.8	&70.2 &66.7 &0.4	&32.0	&24.8	&53.3 &57.6\\
        & \textsc{\qquad\quad~~\texttt{Base SFT}} &6.2 & 36.1 & 29.9 & 70.5 & 66.2 & 0.3 & 31.5 & 24.4 & 52.6 & 57.5\\
        & \textsc{\qquad\quad~~\texttt{\grogu}}     &7.0 & 37.7 & 31.2 & 73.1 & \textbf{70.1} & 0.4 & 33.0 & 25.6 & 55.0 & \textbf{60.1}\\
      
        \bottomrule 
    \end{tabular*}
    \end{threeparttable}
    \caption{Downstream LLM performance on TopiOCQA and QReCC. We calculate the word overlap metrics f1\textsuperscript{*}, prec\textsuperscript{*}, recall\textsuperscript{*} based on the method from \citet{adlakhaTopiOCQAOpendomainConversational2022a}. A test instance (a query) may have more than one gold answers. For each model answer, we compute f1, precision, and recall with respect to every gold answer. We then take the mean of the maximum f1, precision, and recall values for each answer.
    }
    \vspace{-2mm}
    \label{table:reader}
\end{table*}
As presented in \autoref{table:reader}, we observe that our method is able to consistently provide statistically significant improvements in the downstream LLM accuracy.\footnote{The previous works we used in retrieval evaluation do not report the downstream LLM performance, so we are only comparing the untrained models vs. the models we trained.} For example, \grogu\ trained models are able to improve the accuracy of Qwen from 48.8\% (\texttt{Original}) to 56.2\% for TopiOCQA and from 51.4\% to 60.5\% for QReCC when using a BM25 retriever. We observe similar outcomes for the Phi model. When using the ANCE retriever the improvements are smaller in magnitude, as they start from a higher baseline, e.g., Phi improves from 66.7\% to 70.1\% in TopiOCQA. Notably, we notice that our method is improving also with respect to the \texttt{BaseSFT} approach, with gains around 4-8 points in the BM25 setting and .5-4 points with ANCE. The magnitude of changes is smaller than those with the retrieval results, likely a result of the accuracy metric underestimating the actual performance in free-form QA. Yet, they still show that our annotation-free method can provide sizable improvements in generation accuracy.

\section{Additional Implementation Details}
\label{append:implementation}
\paragraph{Grounding Utility Hyperparameters} We tune the $\alpha$ and $K$ hyperparameters for KeyEntropy with the test from \autoref{sec:golddoctest}. We tune them on a small validation set (100 questions) from the train split of the SQuAD dataset~\citep{rajpurkar2016squad100000questionsmachine}, which is disjoint with any of the data we use in this paper. For $\alpha$, We search over a range between 0 and 0.5 with a step size of 0.05. For $K$, we search over a range between 0 and 1 with a step size of 0.1. We find that KeyEntropy is not very sensitive to the choice of $\alpha$ and $K$, and a small $\alpha$ generally works well. We use $\alpha=0.05$ and $K=0.1$ for both Qwen and Phi in our paper.
  
\paragraph{Training Data}
We randomly sample 15555 queries from RetPO's raw \topiocqa dataset and 13238 queries for its raw \qrecc dataset. These are about half the size of the data RetPO used for their data preparation and training pipeline. We choose these dataset sizes to save training compute. We take the 25 GPT-4 generated rewrites for each query from RetPO. There are some queries that RetPO only generated 15 queries, we followed their procedure and generated the remaining 10 queries using GPT-4. For our main experiments, we use the same LLM for producing \grogu and query-rewriting and use the same retriever for training and evaluation. For experiments with larger models, we use a smaller model as the query-rewriter and use the larger models to assign \grogu scores (further explained in \aautoref{append:large_model_training}). 

\paragraph{Utility Scoring} The downstream LLM assigns the \grogu score by taking the top 10 documents retrieved by the rewritten query. This corresponds to how the downstream LLM would use the documents in a simple RAG pipeline, where no reranker is involved and the downstream LLM just takes the top 10 documents.

Though not a focus of this paper, this approach can be easily extended to optimizing a rewriter while assuming access to a common reranker that re-ranks the top 100 retrieved documents and sends K documents to the downstream LLM (e.g., 5 or 10 documents). In this case, we would utilize the GroGU of the top-K documents after the reranker as the training signal. In either case, the way we use GroGU perfectly matches how the downstream LLM would use the documents. The LLM would see the same K documents we measure the GroGU for, which is a feature of our method instead of a compromise. Because this is an initial paper to propose the new paradigm of using reference-free metrics for RAG, we focus on the simpler setup without a reranker, which already strongly supports our paradigm with the significant improvements in retrieval performance (e.g., Recall@10, 20, and 100). We expect the second setup with reranker to produce similar results given that our method also substantially improves Recall@100, which would directly affect the quality of the documents after reranking.

\paragraph{Hyperparameters for Training}
We use the same values from RetPO~\citep{yoonAskOptimalQuestions2025} for all hyperparameters, except for the learning rates. We use 2.0e-5 for SFT warmup following~\citep{yoonAskOptimalQuestions2025}. We search over the learning rates $\{5.0e-7, 5.0e-6, 5.0e-5\}$ for DPO training. \autoref{tab:learning_rates} present the full table of learning rates for DPO training. Similar to RetPO, we train our models for up to 3 epochs for SFT warmup and 3 epochs for DPO. For our SFT baseline (Base SFT), we train it for up to 6 epochs. We take the checkpoint with the smallest validation loss. A single experiment run (SFT+DPO) takes about 2.5 hours with 8 A100 80GB GPUs.
\begin{table}[t]
\centering
\begin{tabular}{@{}lcc@{}}
\toprule
\multicolumn{3}{c}{\textbf{Phi}} \\
\cmidrule(lr){1-3}
 & \textbf{\topiocqa} & \textbf{\qrecc} \\
BM25 & 5e-6 & 5e-7 \\
ANCE & 5e-7 & 5e-6 \\
\cmidrule(lr){1-3}
\multicolumn{3}{c}{\textbf{Qwen}} \\
\cmidrule(lr){1-3}
 & \textbf{\topiocqa} & \textbf{\qrecc} \\
BM25 & 5e-6 & 5e-6 \\
ANCE & 5e-7 & 5e-6 \\
\bottomrule
\end{tabular}
\caption{Learning rates for different experiment scenarios.}
\label{tab:learning_rates}
\end{table}

\paragraph{Statistical Test} We use the sign test for all statistical testing. We use p<0.05 as criterion for statistical significance unless otherwise specified.
\paragraph{Softwares Used}

\begin{itemize}
    \item transformers==4.8.2 (BSD-3 license)
    \item pyserini==0.22.1 (Apache-2.0 license)
    \item torch==2.7.1 (BSD-3 license)
    \item Open-sourced codebase of \citet{cuconasuPowerNoiseRedefining2024} \url{https://github.com/florin-git/The-Power-of-Noise}
    \item Open-sourced codebase of \citet{yoonAskOptimalQuestions2025}. \url{https://github.com/dmis-lab/RetPO?tab=readme-ov-file}
    \item GPT-4 (gpt-4-0125-preview) and GPT-4o-mini (gpt-4o-mini-2024-07-18) \url{https://platform.openai.com/docs/models}
\end{itemize}

AI assistance: we used GPT5 for questions about latex commands during paper writing.
Our use of all softwares is consistent with their intended use.

\section{Additional Results With Larger LLMs}
\label{append:large_model_results}
We present additional results on 32B LLMs, which largely follow the trends of our main experiments with the 7B and 14B models. The models tested are Qwen-2.5-32B-Instruct and Olmo-3.1-32B-Instruct. We choose these models for the same consideration explained in \aautoref{append:model_choice}.
\begin{table*}[]
\centering
\begin{tabular}{llcccc|r}
\hline
& & \multicolumn{2}{c}{\textbf{NQ}} & \multicolumn{2}{c}{\textbf{SQuAD}} & \\
\cmidrule(lr){3-4} \cmidrule(lr){5-6}
\textbf{Model} & \textbf{Metric} & Gold/ & Gold/ & Gold/ & Gold/ & \textbf{Avg} \\
& & Dist & Rand & Dist & Rand & \\
\hline
\multirow{4}{*}{\textbf{Qwen}} 
& PPL & 77.2 & 79.3 & 81.8 & 87.9 & 81.6 \\
& KeyPPL & 79.0 & 82.8 & 82.4 & 89.9 & 83.5 \\
& Entropy & 79.8 & 82.9 & \textbf{85.3} & 90.9 & 84.7 \\
& KeyEntropy & \textbf{79.9} & \textbf{84.0} & 84.3 & \textbf{91.7} & \textbf{85.0} \\
\hline
\multirow{4}{*}{\textbf{Olmo}}
& PPL & 72.0 & 75.7 & 80.1 & 85.9 & 78.4 \\
& KeyPPL & 71.6 & 77.6 & 80.0 & 87.0 & 79.1 \\
& Entropy & \textbf{76.6} & 81.3 & 84.8 & 90.3 & 83.3 \\
& KeyEntropy & 75.9 & \textbf{83.2} & \textbf{85.5} & \textbf{91.4} & \textbf{84.0} \\
\hline
\end{tabular}
\caption{32B models' win rates of gold vs. distractor/random with different metrics on Natural Questions (NQ) and SQuAD.}
\label{table:gtwinrate_large}
\end{table*}
\subsection{Inference-only Experiments}
\autoref{table:gtwinrate_large} reports the win rates of the gold
document against the distractor and the random document, as defined in \autoref{sec:golddoctest} ``\golddoctest''. We observe that all metrics give non-trivial performance in identifying the gold documents, much greater than random guess (50\%). The entropy-based metrics still overall achieve better performance than their perplexity-based counterparts. KeyEntropy also still achieves the best performance on average, though Entropy's performance is close to KeyEntropy's. 

\begin{table*}[h]
\centering
\begin{tabular}{llccc}
\hline
\textbf{Model} & \textbf{Metric} & $\tau$ & \textbf{Acc (\%)} & \textbf{F1 (\%)} \\
\hline
\multirow{4}{*}{\textbf{Qwen}} 
& PPL & 0.189 & 59.4 & 53.4 \\
& KeyPPL & 0.246 & 62.3 & 54.4 \\
& Entropy & 0.246 & 62.3 & 55.3 \\
& KeyEntropy & \textbf{0.288} & \textbf{64.4} & \textbf{57.5} \\
\hline
\multirow{4}{*}{\textbf{Olmo}}
& PPL & 0.068 & 53.4 & 52.8 \\
& KeyPPL & 0.107 & 55.4 & 54.9 \\
& Entropy & 0.160 & 58.0 & 57.1 \\
& KeyEntropy & \textbf{0.186} & \textbf{59.3} & \textbf{58.7} \\
\hline
\end{tabular}
\caption{Simple classifier performance and the correlation between utility and accuracy ($\tau$) for 32B models.}
\label{table:correl_w_acc_large}
\end{table*}
\autoref{table:correl_w_acc_large} is also consistent with the findings in \autoref{sec:randdoctest}, where all the utility metrics show a positive correlation with the generation performance, with KeyEntropy significantly outperforming all the other formulations. 

Finally, \autoref{table:same_doc_diff_utils_large} shows the results from the \samedoctest experiments (defined in \autoref{sec:samedoctest}). It shows that both the 32B models still display some level of order effects bias and our metric can help identify the order of documents that's most useful to a model, increasing Olmo's performance from 63.8 (random baseline) to 66.2 and Qwen's performance from 83.8 (random baseline) to 84.7. We notice that the documents assigned higher utility by Qwen (i.e., $\mathcal{D}_{\text{Qwen}}$) also leads to a slightly higher accuracy for Olmo (64.9), though they are still not as effective as those assigned high utility by Olmo itself ($\mathcal{D}_{\text{Olmo}}$), which gives an accuracy of 66.2.

\begin{table*}[]
\centering
\begin{tabular}{lccccc}
\hline
& \multicolumn{4}{c}{\textbf{Metric}} & \\
\cmidrule(lr){2-5}
& PPL & KeyPPL & Entropy & KeyEntropy & Random \\
\hline
Olmo($\mathcal{D}_{\text{Olmo}}$) & 64.4 & 65.2 & 65.7 & \textbf{66.2} & \multirow{2}{*}{63.8} \\
Olmo($\mathcal{D}_{\text{Qwen}}$) & 65.1 & 64.8 & 64.8 & 64.9 & \\
\hline
Qwen($\mathcal{D}_{\text{Qwen}}$) & 84.0 & 84.2 & 84.5 & \textbf{84.7} & \multirow{2}{*}{83.8} \\
Qwen($\mathcal{D}_{\text{Olmo}}$) & 83.1 & 84.3 & 83.1 & 83.5 & \\
\hline
\end{tabular}
\caption{Generation Accuracy for different contexts scenario for 32B models. We present the random baselines in the last column, which correspond to $\text{Olmo}(\mathcal{D}_{\text{rand}})$ and $\text{Qwen}(\mathcal{D}_{\text{rand}})$.}
\label{table:same_doc_diff_utils_large}
\end{table*}

\subsection{Training Experiments}
\label{append:large_model_training}
\paragraph{Setup} Due to the constraints of our computational resources, our experiments with larger models do not use every combination of experiment conditions that we used in our main experiments. Nonetheless, the experiments we conducted still demonstrate that our metric generalizes well to larger models. Specifically, we use the Qwen-2.5-7B-instruct model as the backbone of the query-rewriter and use a larger 32B model as the downstream LLM that assigns the utility scores. It's different from our main experiments' setting, where we used the same small model as both the query-rewriter and the downstream LLM. This new setting actually represents a realistic scenario where the query-rewriter is a relatively small model and the downstream LLM is larger. In our previous main experiments, we used the same retriever for training data construction and evaluation. For the large model experiments, we only use the BM25 retriever when constructing the training datasets but we also evaluate the final trained rewriter with ANCE. 

\paragraph{Results} As presented in \autoref{table:main_large}, we see consistent, substantial improvement over the off-the-shelf model and Base SFT, a trend similar to our main results in \autoref{table:main}. Additionally, even though we only used BM25 as the retriever to construct the training data, we still see large improvements when evaluating with the ANCE retriever. This means that the query-rewriter trained with one retriever generalizes well to another retriever, which is consistent with the findings in the original RetPO paper~\citep{yoonAskOptimalQuestions2025}.
\begin{table*}[h]
    \footnotesize
    \centering
    \begin{threeparttable}
    \renewcommand{\arraystretch}{0.6}

    \begin{tabular*}{0.93\textwidth}{cl|cccc|cccc}
        \toprule
        & &   \multicolumn{4}{c}{\textbf{TopiOCQA}} &  \multicolumn{4}{c}{\textbf{QReCC}}  \\ 
        \textbf{Type} & \textbf{Method} & \textbf{MRR} & \textbf{R@10}  & \textbf{R@20} & \textbf{R@100} & \textbf{MRR} & \textbf{R@10} & \textbf{R@20} & \textbf{R@100} \\
        \midrule 
        \multirow{9}{*}{\rotatebox[origin=c]{90}{\textbf{Sparse (BM25)}}} 
        
        & \textsc{\texttt{Before Post-training}}  &14.6 & 27.8 & 35.5 & 52.7 & 27.7 & 43.6 & 50.7 & 65.4 \\
        & \textsc{\qquad\quad~~\texttt{Base SFT}} &17.4 & 32.8 & 41.6 & 60.5  &32.4 & 51.4 & 60.1 & 75.5   \\
        & \textsc{\qquad\quad~~\texttt{\grogu (Qwen-32b)}} &27.2	&49.8	&59.3	&74.5 &47.0	&67.4	&75.2	&85.6\\
        & \textsc{\qquad\quad~~\texttt{\grogu (Olmo-32b)}} &26.8	&48.1	&57.4	&74.3	&45.1	&65.9	&73.6	&84.8\\
        \cmidrule(lr){2-10}
        & RetPO   & \textbf{28.3} & 48.3 & -  & \textbf{73.1} & 50.0 & 69.5 & - & 89.5 \\
        & ConvGQR & 12.4 & 23.8 & -    & 45.6 & 44.1 & 64.4 & - & 88.0 \\
        & IterCQR & 16.5 & 29.3 & -    & 54.1 & 46.7 & 64.4 & - & 85.5 \\
        & ADACQR  & \textbf{28.3} & \textbf{48.9} & -    & 71.2 & \textbf{55.1} & \textbf{76.5} & -    & \textbf{93.7}\\
        \midrule 
        \multirow{9}{*}{\rotatebox[origin=c]{90}{\textbf{Dense (ANCE)}}} 
        & \textsc{\texttt{Before Post-training}}   &14.6 & 27.8 & 35.5 & 52.7 & 27.7 & 43.6 & 50.7 & 65.4 \\
        & \textsc{\qquad\quad~~\texttt{Base SFT}} &17.4 & 32.8 & 41.6 & 60.5  &32.4 & 51.4 & 60.1 & 75.5   \\
        & \textsc{\qquad\quad~~\texttt{\grogu (Qwen-32b)}} &41.1 &63.3 &70.8  &81.7	&40.7	&60.8	&67.4	&76.5\\
        & \textsc{\qquad\quad~~\texttt{\grogu (Olmo-32b)}} &39.6  &63.0	&69.9	&80.1 &40.6	&60.8	&67.5	&76.7\\
        \cmidrule(lr){2-10} 
        & RetPO   & 30.0 & 49.6 & - & 68.7 & 44.0 & 66.7 & - & \textbf{84.6} \\ 
        & ConvGQR & 25.6 & 41.8 & - & 58.8 & 42.0 & 63.5 & - & 81.8 \\
        & IterCQR & 26.3 & 42.6 & - & 62.0 & 42.9 & 65.5 & - & 84.1 \\
        & ADACQR  & \textbf{38.5} & \textbf{58.4} & - & \textbf{75.0} & \textbf{45.8} & \textbf{67.3} & - & 83.8 \\
        \bottomrule 

    \end{tabular*}
    \end{threeparttable}
    \caption{Sparse and dense retriever performance on TopiOCQA and QReCC. \texttt{\grogu (Qwen-32b)} and \texttt{\grogu (Olmo-32b)} represent experiments where the preference data are produced by the utility scores assigned by the 32B models, showing that our metric generalizes well to larger models. Baseline models' performance scores are those reported in their original papers, which do not report Recall@20.}
    \vspace{-2mm}
    \label{table:main_large}
\end{table*}

\section{Related But Not Comparable Methods}
\label{append:not_comparable_methods}
\begin{table*}[h]
    \footnotesize
    \centering
    \begin{threeparttable}
    \renewcommand{\arraystretch}{0.6}

    \begin{tabular*}{0.93\textwidth}{cl|cccc|cccc}
        \toprule
        & &   \multicolumn{4}{c}{\textbf{TopiOCQA}} &  \multicolumn{4}{c}{\textbf{QReCC}}  \\ 
        \textbf{Type} & \textbf{Method} & \textbf{MRR} & \textbf{R@10}  & \textbf{R@20} & \textbf{R@100} & \textbf{MRR} & \textbf{R@10} & \textbf{R@20} & \textbf{R@100} \\
        \midrule 
        \multirow{9}{*}{\rotatebox[origin=c]{90}{\textbf{BM25}}} 
        
        & \textsc{\texttt{\grogu} - Qwen} &25.7 & 46.1 & 54.7 & 72.0 &45.9 & 67.2 & 74.9 & 85.6 \\
        & \textsc{\texttt{\grogu} - Phi}  &27.2 & \textbf{48.8} & \textbf{58.6} &\textbf{74.8} & 45.8 & 66.2 & 73.7 & 84.3\\
        & RetPO   & \textbf{28.3} & 48.3 & -  & 73.1 & \textbf{50.0} & \textbf{69.5} & - & \textbf{89.5} \\
        
        \cmidrule(lr){2-10}
        & Query2Doc - Qwen & 16.7&	29.2&	35.0&	47.7 & 38.5	& 54.4& 60.7	& 70.7\\
        & Query2Doc - Phi &\textbf{24.5}	&42.2	&\textbf{49.9}	&66.2 &\textbf{48.6}	&67.6	&\textbf{73.8}	&82.5	\\
        & ConvSearch-R1 & - & \textbf{59.6} & - & \textbf{80.1} & - & \textbf{77.2} & - & \textbf{89.0}\\
        \midrule 
        \multirow{9}{*}{\rotatebox[origin=c]{90}{\textbf{ANCE}}} 
        & \textsc{\texttt{\grogu} - Qwen} &37.8 & \textbf{61.7} & \textbf{68.7} & \textbf{79.8} &40.2 & 60.8 & 67.5 & 76.8\\
        & \textsc{\texttt{\grogu} - Phi}      &\textbf{38.1} & 61.1 & 68.3 & 79.6 & 39.4 & 59.8 & 67.1 & 76.3\\
        & RetPO  & 30.0 & 49.6 & - & 68.7 & \textbf{44.0} & \textbf{66.7} & - & \textbf{84.6} \\ 
        
        \cmidrule(lr){2-10} 
        & Query2Doc - Qwen & 35.7	&56.2	&63.2	&74.5 &40.5	&59.5	&66.3 &74.6\\
        & Query2Doc - Phi & \textbf{36.5}	&58.5	&\textbf{65.6}	&77.6   &\textbf{43.6} &64.1	&\textbf{70.9}	&78.7 \\
        & ConvSearch-R1 & - & \textbf{72.0} & - & \textbf{86.3} & - & \textbf{70.6} & - & \textbf{82.8}\\
        \midrule
        & Promptriever & 7.4 & 12.8 & 15.4 & 22.6 & 11.4 & 19.9 & 24.9 & 32.1\\

        \bottomrule 
    \end{tabular*}
    \end{threeparttable}
    \caption{Performance of other not comparable methods. ``-'' indicates values not reported in the method's paper.}
    \vspace{-2mm}
    \label{table:not_comparable_baseline}
\end{table*}

We report the performance of the methods that are not comparable with ours in \autoref{table:not_comparable_baseline}. Promptriever~\citep{weller2024promptriever} is based on a versatile LLM and designed to understand any input instruction/text to retrieve relevant documents. One may expect it to understand the conversational context and the instruction to find documents that answer the last question without any training. However, we empirically find that it performs much worse than our method or the comparable baseline RetPO, which are finetuned for the conversational query-rewriting task.

Query2Doc~\citep{wang-etal-2023-query2doc} uses few-shot examples to prompt an LLM to write a pseudo-document that answers the given question. Given the LLM’s parametric knowledge of what may be relevant and the longer query (a passage rather than a question), one may expect this method to be applicable to conversational query-rewriting. We find that it still performs worse than our method or other baselines, despite incurring substantially higher generation costs. It only performs on par with or better than our method in a small number of scenarios (such as with Phi-4 on the QReCC dataset). 

ConvSearch-R1 requires gold labels and is thus a supervised method, unlike ours. It also uses reasoning chains and online RL training, which require substantially more compute for both training and inference than our method and other baselines. Because of this, it is not surprising that ConvSearch-R1 overall outperforms the comparable baseline, RetPO. However, the high demand for inference-time compute makes it less suitable for practical applications, where the retrieval process needs to be done efficiently to minimize latency and compute cost. Additionally, ConvSearch-R1's benefits become notably smaller at Recall@100, so in practical retrieval pipelines where a reranker would be applied to the top-100 documents, the performance gap would have even less impact on the final grounding documents' quality.

Finally, we expect that our metric can replace ConvSearch-R1's training reward that is based on gold passage labels, and leave this study to future work. For this paper, we choose RetPO as the framework to test our metric because ConvSearch-R1 requires too much resource to train (beyond our current capacity) and its disadvantage in efficient practical applications due to the high inference cost.

\end{document}